\pdfoutput=1
\documentclass[11pt]{article}

\usepackage{acl}

\usepackage{times}
\usepackage{latexsym}

\usepackage[T1]{fontenc}
\usepackage[utf8]{inputenc}

\usepackage{microtype}

\usepackage{inconsolata}

\usepackage{graphicx}

\usepackage{amsmath} 

\usepackage{alterqcm}
\title{Is There a Case for Conversation Optimized Tokenizers in Large Language Models?}

\author{ R. Ferrando, J. Conde. G. Martínez and P. Reviriego, \\
	ETSI de Telecomunicación\\
	Universidad Politécnica de Madrid\\
	28040 Madrid, Spain \\
}

\begin{document}
\maketitle
\begin{abstract}
The computational and energy costs of Large Language Models (LLMs) have increased exponentially driven by the growing model sizes and the massive adoption of LLMs by hundreds of millions of users. The unit cost of an LLM is the computation of a token. Therefore, the tokenizer plays an important role in the efficiency of a model, and they are carefully optimized to minimize the number of tokens for the text in their training corpus. One of the most popular applications of LLMs are chatbots that interact with users. A key observation is that, for those chatbots, what is important is the performance of the tokenizer in the user text input and the chatbot responses. Those are most likely different from the text in the training corpus. So, a question that immediately arises is whether there is a potential benefit in optimizing tokenizers for chatbot conversations. In this paper, this idea is explored for different tokenizers by using a publicly available corpus of chatbot conversations to redesign their vocabularies and evaluate their performance in this domain. The results show that conversation-optimized tokenizers consistently reduce the number of tokens in chatbot dialogues, which can lead to meaningful energy savings, in the range of 5\% to 10\% while having minimal or even slightly positive impact on tokenization efficiency for the original training corpus.

\end{abstract}

\section{Introduction}

The exponential development and adoption of generative AI requires a large and growing amount of computing and energy resources, creating concerns about its sustainability \cite{bigtechenergy}. This has placed the energy efficiency of LLMs at the center of the different approaches to reduce the energy dissipation of AI \cite{energyinference}. In most models, the energy consumed is proportional to the number of tokens in the input and output text \cite{energytoken}. Therefore, given a text, a tokenizer that reduces the number of tokens would improve energy efficiency. 

The design of tokenizers given their importance has been widely studied \cite{mielke2021between}. They are typically optimized to reduce the number of tokens needed for their training corpus by assigning tokens to the most common words or sub-word units using different algorithms \cite{wang2019neuralmachinetranslationbytelevel}. 

A key observation is that, in many cases, LLMs are used to power chatbots that interact with users such as ChatGPT. In that application, the input text to the LLM and the generated text will be significantly different from their training corpus, which includes books, texts from the Internet, or documents and is not focused on conversations. Therefore, existing tokenizers are not optimum for the most common application of LLMs. The question that arises immediately is how large the benefit of optimizing the tokenizers for conversational applications would be. 

This paper explores the potential benefits of optimizing tokenizers for conversational text. In more detail, a representative group of tokenizers are re-trained using a publicly available dataset of chatbot conversations and their performance is compared to the original tokenizers on another set of conversations. The results show that in all cases performance improves and for some tokenizers the reduction in the number of tokens reaches more than 10\%. This means that relevant energy savings can be obtained when generating text using conversation-optimized tokenizers. However, the results are preliminary and need to be confirmed using other conversational datasets. Additionally, the tokenizer also has an impact on the training phase of LLMs and thus the impact of optimizing the tokenizers for the inference phase using conversations may induce an additional cost in the training phase which also has to be evaluated. Finally, modifying the tokenizer may also affect the LLM performance for different tasks, so this needs to be evaluated.

The rest of the work is organized as follows: in section \ref{sec:methodology} the methodology used in our analysis is presented, followed by the results in section \ref{sec:ResultsAnalysis} and a discussion of their implications in section \ref{sec:discussion}. The paper ends with the conclusion in section \ref{sec:Conclusion}.

\section{Methodology}
\label{sec:methodology}

To evaluate the potential benefits of optimizing the tokenizers for conversations, we first select a representative group of tokenizers and two relevant datasets: one of chatbot conversations and the other an LLM training corpus. Then we first check that existing tokenizers have worse performance on conversations than on a training corpus to confirm our initial hypothesis. In a second experiment, we optimize the tokenizers for conversations and evaluate the reduction obtained in the number of tokens. Finally, we also evaluate the performance of conversation-optimized tokenizers in the training corpus. 

\subsection{Tokenizers}

Tokenization is the division of the input text of an LLM into predefined linguistic units called tokens, which can be characters, word fragments, or whole words. Each tokenizer has its own tokenization rules and token vocabulary \cite{mielke2021between}. The vocabulary is designed using a statistical process that seeks to identify the best subwords to represent a text corpus, based on the frequency and distribution of text fragments \cite{tokenizers_medium}. The exact way tokens are selected depends on the tokenization algorithm used. The process is completely deterministic: training with the same algorithm and the same corpus always produces the same vocabulary \cite{hf_tokenizers}. 

To achieve efficient text compression, according to the principles of information theory, it is essential to construct an optimized vocabulary so that the most frequent words in the training corpus are represented with fewer tokens, thereby minimizing the total length of tokenized sequences. Several algorithms effectively implement this approach. Among the most used are Byte Pair Encoding (BPE), WordPiece, and Unigram. For a given vocabulary size, each uses a specific methodology to segment and prioritize lexical units and add them to the token vocabulary. 

For evaluation, a group of tokenizers used in popular LLMs has been selected. The models considered are summarized in Table \ref{tab:models} and include both large proprietary models from OpenAI and also open-weights models from different companies or open-source initiatives. The main features of their tokenizers are shown in Table \ref{tab:vocabtokenizers}. This group provides a representative set of the tokenizers used in LLMs at the time of writing this paper.

\begin{table*}
\centering
{\small
\begin{tabular}{|l|l|p{9cm}|}
\hline
\textbf{Model} & \textbf{Organization} & \textbf{Description} \\ \hline
GPT-4 & OpenAI & Reference model in chatbot applications, with the highest number of active users to date \cite{OpenAI2023GPT4, demandsagechatgpt}. \\ \hline
GPT-4o & OpenAI & Latest and fastest model, optimized for performance. Available for free but with usage limits \cite{openai_gpt4o_2024}. \\ \hline
DeepSeek-R1 & DeepSeek & Model optimized for reasoning tasks and widely adopted \cite{deepseek-llm}. \\ \hline
LLaMA-3.1-8B & Meta & Lightweight and fast model, a prominent example of the latest generation of open models \cite{Llama3.1}. \\ \hline
Gemma-2-9b & Google & Lightweight and open alternative with good performance and focus on safety and efficiency \cite{Schmid2024Gemma2}.\\ \hline
Mistral-7B-v0.1 & Mistral AI & High-capacity model for its small size \cite{mistral2}. \\ \hline
BLOOM & BigScience & Open multilingual model, designed with a collaborative and responsible approach \cite{workshop2023bloom176bparameteropenaccessmultilingual}. \\ \hline
Phi-4 & Microsoft & Lightweight model oriented to reasoning tasks, recently presented by Microsoft \cite{microsoft2024phi4}. \\ \hline
\end{tabular}
}
\caption{Selected models for evaluation of their tokenizers}
\label{tab:models}
\end{table*}

\begin{table*}
\centering
{\small
\begin{tabular}{|l|p{3cm}|p{9cm}|}
\hline
\textbf{Model} & \textbf{Vocabulary Size (tokens)} & \textbf{Tokenization Features} \\ \hline
GPT-4 & $\approx 100,000 $ & Byte-Level BPE algorithm. Tokenizer available in \texttt{tiktoken} \cite{openai2023tiktoken}. The token vocabulary is called "cl100k\_base", but it's not accessible, as with the tokenizer configuration details. Also used by GPT-3.5 and GPT-3.5 turbo models. \\ \hline
GPT-4o & $\approx 200,000 $ & Optimized version of the GPT-4 tokenizer, with an expanded vocabulary called "o200k\_base". Used in reasoning models o1 and o3 \cite{openai2023tiktoken}. \\ \hline
DeepSeek-R1 & 128,815 & Byte-Level BPE algorithm. In Hugging Face, it is implemented using the same structure as the LLaMA tokenizer. Also used by DeepSeek-V3.  \\ \hline
LLaMA-3.1-8B & 128,000 & Byte-Level BPE. The same tokenizer is used across all models in the LLaMA 3 series, including versions 3.1, 3.2, and 3.3. \\ \hline
Gemma-2-9b & 256,000 & Character-level BPE, including a byte-fallback mechanism to avoid out-of-vocabulary tokens. It was trained using SentencePiece \cite{kudo-2018-subword}. All models in the Gemma1, Gemma2, TXGemma, and CodeGemma families share the same tokenizer. \\ \hline
Mistral-7B-v0.1 & 32,000 & V3 tokenizer \cite{mistral_tokenization}, shared by the following sets of models: open-mixtral-8x22b, mistral-large-latest, mistral-small-latest, open-mistral-7b. Character-level BPE with byte-fallback mechanism, implemented using SentencePiece.\\ \hline
BLOOM & 250,680 & Byte-Level BPE. It was developed with a multilingual focus. It was trained on a subset of the ROOTS dataset, preserving the same language distribution as the model training data. The vocabulary size was chosen to ensure suitable fertility across languages and is a multiple of 128 to enhance GPU efficiency and parallelism \cite{workshop2023bloom176bparameteropenaccessmultilingual}.\\ \hline
Phi-4 & 100,352 & Implemented with the same scheme as OpenAI's Open Source model, GPT-2. It has a very similar or identical token vocabulary size to GPT-4. It has been verified that tokenizing several sequences with both the Phi-4 and GPT-4 tokenizers produces almost the same token sequences, indicating both tokenizers are almost identical.  \\ \hline
\end{tabular}
}
\caption{Comparison of Models' Vocabulary and Tokenization Features}
\label{tab:vocabtokenizers}
\end{table*}

\subsection{Datasets}

Two text corpora are needed for the evaluation, one consisting of chatbot conversations and another that corresponds to text used to train tokenizers. The corpora used are briefly described in the following subsections.

\subsubsection{Chatbot Conversation Corpus: LMSYS Chat 1M}

The LMSYS Chat 1M dataset \cite{zheng2023lmsyschat1m}, available on Hugging Face \cite{huggingface}, contains one million conversations from anonymous users with 25 different LLMs in multiple languages \cite{zheng2023lmsyschat1mpaper}. Each dataset entry includes a conversation identifier, the LLM generating the response, the conversation text, the detected language, and a moderation tag provided by OpenAI’s API. 

This corpus is ideal for evaluating tokenization in the conversational context, as it consists of real conversations with chatbots. Therefore, it is representative of the type of text processed and generated by language models in conversational applications.

\subsubsection{LLM Training Corpus: C4 by AllenAI}
There are numerous open-access datasets for training LLMs. Some of the most widely used include Common Crawl \cite{commoncrawl}, The Pile \cite{gao2020pile}, and OpenWebText \cite{Gokaslan2019OpenWeb}. All contain vast amounts of text, ranging from tens to hundreds of gigabytes. For example, The Pile is approximately 800 GB.

For tokenization evaluation purposes, using such a large corpus is unnecessary and computationally costly. Therefore, one of the criteria for selecting the most suitable training text corpus has been the ability to easily generate a small representative subset. The accessibility of the dataset through platforms like Hugging Face, which allow easy and efficient download and data manipulation, has also been considered. According to these criteria, the chosen dataset is the Colossal Clean Crawled Corpus (C4) \cite{allenai2020c4}, developed by the Allen Institute for AI (AllenAI) \cite{allenai2024website}. It is a clean version of the Common Crawl, removing duplicate content, low-information pages, spam, and other unwanted elements. The result is an extensive corpus representative of modern web language, widely used in the training and evaluation of language models. Therefore, it is a good representation of the text used to train tokenizers.

\subsection{Evaluation procedure and metrics}

\subsubsection{Procedure}

The procedure used to design conversation-optimized tokenizers is to re-train the tokenizer using the same algorithm and configuration but on a subset of the LMSYS Chat 1M dataset. Then the new tokenizers are evaluated in the rest of the LMSYS Chat 1M dataset. The train and test split was 80\%-20\% and the conversations were randomly assigned to one of the sets. Different strategies are used for the optimization: to use only the user input text, only the chatbot responses, or both to re-train the tokenizer. This is done to explore the differences between the questions that are generated by the users and the responses that are generated by the LLMs.   

\subsubsection{Metrics}

The main performance metric in our evaluation is the number of tokens needed for a given text. When the text is the same and we are comparing the original and conversation-optimized tokenizers the ratio of tokens needed by each tokenizer is used to measure the gain or loss introduced by optimizing the tokenizer. However, when we want to understand the relative performance of a tokenizer on different types of texts, a ratio of tokens can not be used as the sizes of the texts are different. In this case, fertility \cite{rust2021goodtokenizermonolingualperformance} defined as the number of tokens per word in a given text is used, the metric is defined by:   

\begin{equation}
Fertility = \frac{N_{\text{tokens}}}{N_{\text{words}}}
\end{equation}

and values closer to one correspond to a more efficient tokenization of the text.

\section{Results and Analysis}
\label{sec:ResultsAnalysis}

The evaluation is performed in three phases. In the first experiment, we check that current LLM tokenizers perform better on the LLM training corpus than in the conversational dataset. In the second experiment, we design conversation-optimized tokenizers and evaluate the reduction in the number of tokens for the conversational dataset. Finally, in the last experiment, we measured the performance of conversation-optimized tokenizers in the LLM training corpus to understand the impact of optimization for other LLM applications. The results for each of the experiments are presented in the following subsections. The raw data and the scripts to reproduce the experiments are available in a public repository\footnote{\url{https://github.com/RaquelFerrando/conversational_tokenizers.git}}.

\subsection{Experiment 1: Performance of current tokenizers in conversations versus training corpus}

\begin{figure*}[h]
  \centering
  \includegraphics[scale=0.4]{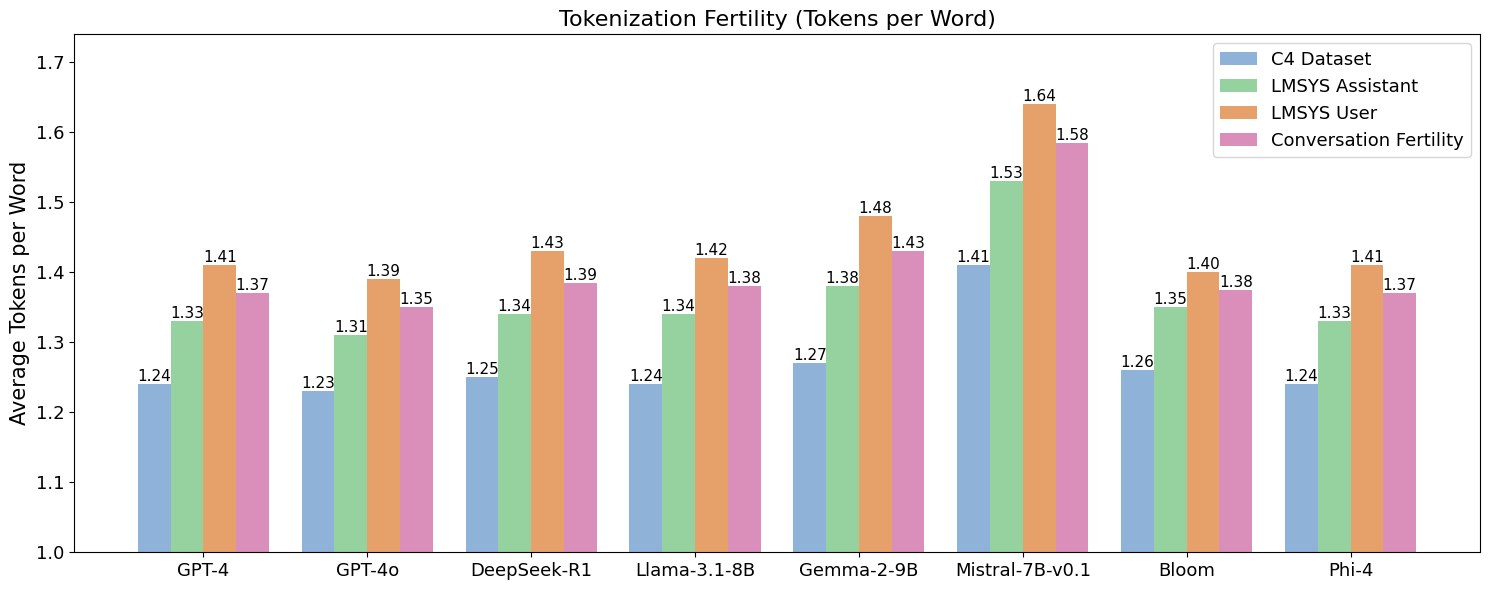}
  \caption{Fertility on the LLM training dataset (C4) and on the conversational dataset (LMSYS). For the conversational dataset fertility is computed on the input (user), output (assistant) and entire conversations.}
  \label{fig:fertility}
\end{figure*}

The results in terms of fertility of the different tokenizers in the LLM training corpus and the chatbot conversational dataset are summarized in Figure \ref{fig:fertility}. To ensure a more reliable comparison, both the training and conversational texts are in English. It can be observed that for all the models the fertility is lower on the training dataset than on the conversations. This confirms our intuition that tokenizers designed with LLM training corpora are not optimum for conversational applications. For conversations, the fertility is also computed on the user inputs only and on the chatbot responses only. It can be seen that the fertility is lower for the chatbot responses than for the user questions, which can be due to the nature of the text in the responses, but also to the fact that LLMs may tend to create text that is aligned with their tokenizers. Finally, the overall fertility for conversations is closer to that of the responses, this is because responses account for the majority of the text in the conversations.   

\subsection{Experiment 2: Performance of conversation-optimized tokenizers: improvements for conversations}

\begin{figure*}[h]
  \centering
  \includegraphics[scale=0.4]{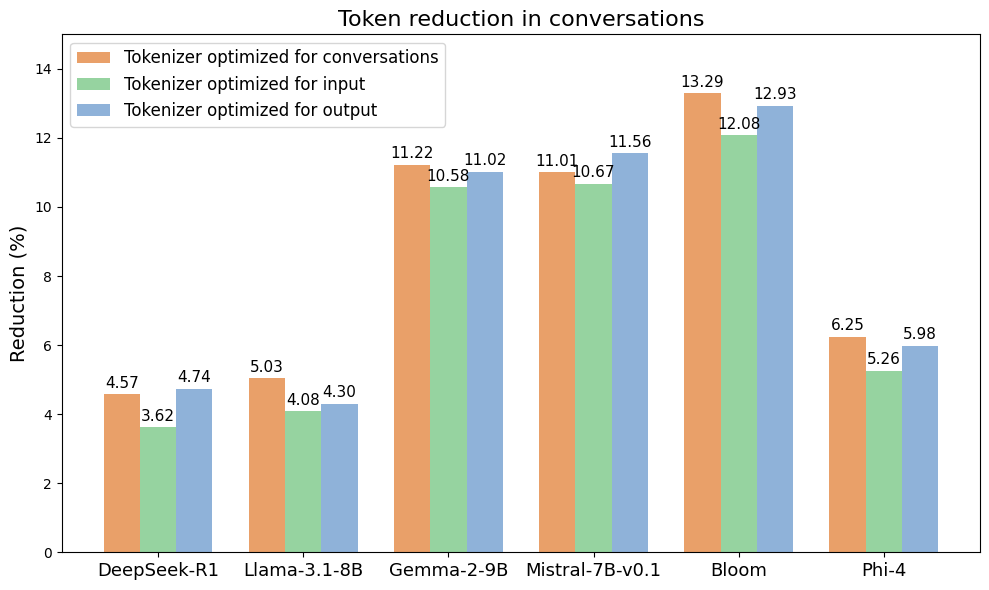}
  \caption{Reduction on the number of tokens for the conversation-optimized tokenizers on the conversational test set (LMSYS).}
  \label{fig:reduction-all}
\end{figure*}

To assess the potential benefits of optimizing LLM tokenizers for conversations, for each model, three optimized tokenizers are constructed using the user inputs only, the chatbot outputs only, and both. Then they are used to tokenize the test set, and the reduction versus the original tokenizers is computed. The results are summarized in Figure \ref{fig:reduction-all}. It can be observed that all three tokenizers reduce the number of tokens but savings are larger when optimizing for the entire conversations or only for the output (which accounts for the majority of the words). Therefore, in order to maximize the savings, it is better to optimize the tokenizer for the entire conversations.

The reductions vary across models. Savings are approximately 5\% for DeepSeek-R1, Llama3-1-8B and Phi-4, while for Gemma-2-9B, Mistral-7B and Bloom they exceed 10\%.

\begin{figure*}[h]
  \centering
  \includegraphics[scale=0.45]{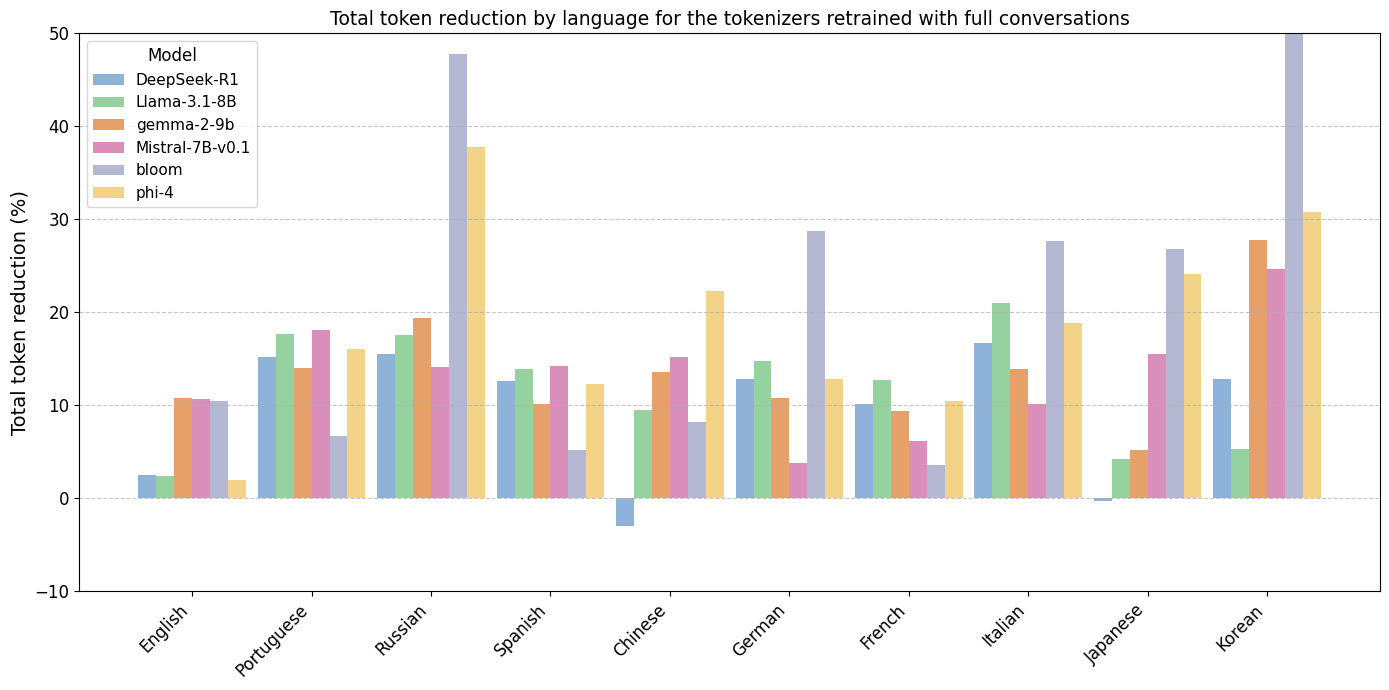}
  \caption{Token reduction achieved by conversation-optimized tokenizers on the top 10 languages in the LMSYS dataset (i.e., languages with over 1,000 conversations in the test corpus).}
  \label{fig:reduction-languages}
\end{figure*}

It is important to note that the test set is multilingual, and the language distribution of the conversations may also influence the observed token reductions. Since tokenization efficiency can vary significantly between languages, the reductions are not uniform across all of them. However, a language-wise analysis shows that the languages that are well represented in the dataset generally benefit from using the conversation-optimized tokenizers, as shown in Figure \ref{fig:reduction-languages}. This is confirmed by the poor performance of DeepSeek in Chinese, for which the number of tokens increases when the tokenizer is optimized for conversations. This is probably due to the training set having a larger fraction of Chinese than the conversational data set that only has 2.4\% of the conversations in Chinese.

% with Mistral-7B achieving over 20\% and Bloom more than 10\%. This can be attributed to ???

\subsection{Experiment 3: Performance of conversation-optimized tokenizers: loss for training corpus}

\begin{figure*}[h]
  \centering
  \includegraphics[scale=0.4]{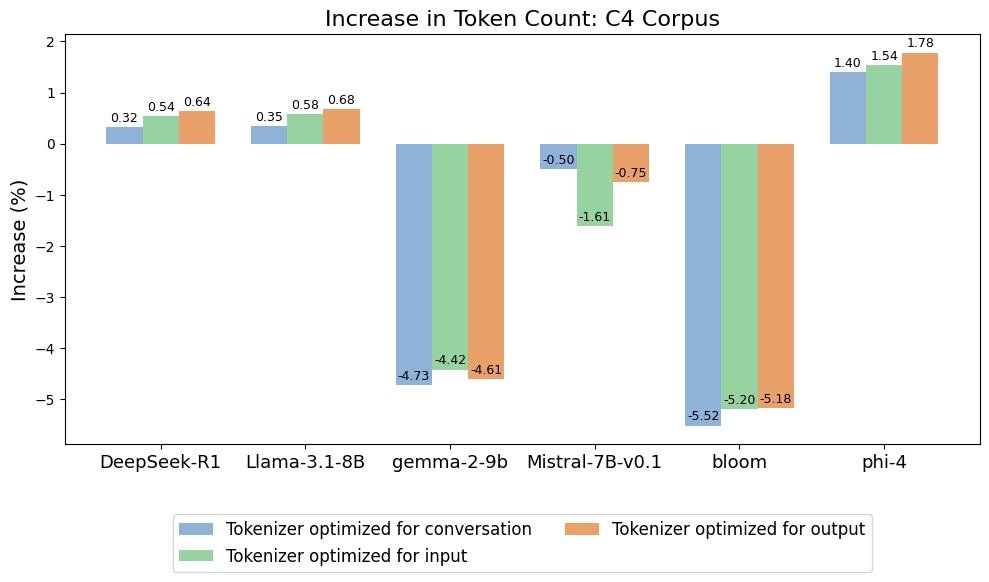}
  \caption{Increase on the number of tokens for the conversation-optimized tokenizers on the LLM training corpus (C4).}
  \label{fig:increase-all}
\end{figure*}

In the last experiment, we ran the tokenizers optimized for the conversational dataset on the LLM training corpus and computed the increase in the number of tokens compared to the original tokenizers. The results are summarized in Figure \ref{fig:increase-all}. 

Surprisingly, three of the models, Mistral-7B, Gemma-2-9B, and Bloom, show a reduction in the number of tokens: approximately 1\% for Mistral-7B and around 5\% for both Gemma-2-9B and Bloom. The fact that tokenizers optimized for conversations also lead to a reduction in token count on the training corpus may suggest that the original tokenizers are sub-optimal. These reductions are, in all cases, smaller than those obtained on the conversational corpus. The same three models were also the ones showing the largest reductions for conversations. This might indicate that the improvements observed in conversations are partly due to the specific optimization for that domain, and partly due to more general inefficiencies in the original tokenizers, even outside of conversational data.

For the rest of the models, DeepSeek-R1, Llama3-1-8B, and Phi-4, the number of tokens is slightly increased, but the values are below 2\%. The best performance is achieved when optimizing the tokenizer for the entire conversations. Overall, it seems that optimizing the tokenizer for conversations does not have a large impact on the performance on the LLM training text.

\section{Discussion}
\label{sec:discussion}

The analysis and results show that current LLM tokenizers are not optimal for chatbot applications  and that the computational and energy costs could be reduced by designing conversation-optimized tokenizers. 

This observation is particularly relevant given the growing importance of energy efficiency in AI. As LLMs are increasingly deployed in real-time conversational settings \cite{demandsagechatbot}, optimizing tokenization for this use case presents an opportunity to reduce resource consumption during inference, which accounts for a significant portion of the long-term energy footprint of AI systems, likely larger than that of training \cite{wordstowatts}.

However, there are several considerations to keep in mind. While optimizing for conversational text may lead to energy savings during inference in chatbot applications, it could slightly increase the cost of training. Moreover, tokenization impacts not only the efficiency but also the effectiveness of training. A tokenizer optimized for conversations might influence the learning process in ways that affect the model’s downstream performance.

Overall, our findings highlight a promising direction for improving the efficiency of LLMs in real-world applications in a simple way. Tokenizer design, often overlooked compared to model architecture or training data, can have a measurable impact on energy consumption. Revisiting this layer with application-specific constraints in mind could unlock a new dimension of energy-aware optimization for language models. Still, such changes should be approached with care, as they may introduce trade-offs that need to be carefully evaluated.

\section{Conclusion and future work} 
\label{sec:Conclusion}
%This paper has explored the potential of optimizing tokenizers for chatbot conversations as a simple strategy to improve the efficiency of Large Language Models (LLMs) during inference. We show that current tokenizers perform suboptimally in conversational contexts, one of the most common real-world use cases of LLMs. By retraining tokenizers on a representative dataset of chatbot dialogues, we observed consistent reductions in the number of tokens required, with savings ranging from 5\% to over 10\% depending on the model. These reductions can translate into meaningful energy savings at inference time.

%Nonetheless, modifying the tokenizer can affect the model's learning dynamics and downstream performance. Therefore, future research should evaluate and check that improvements in efficiency do not come at the expense of model quality or generalization.

%Further analysis of the reductions in conversations and other types of texts. 

This paper has explored the potential of optimizing tokenizers for chatbot conversations as a simple strategy to improve the efficiency of Large Language Models (LLMs) during inference. We show that current tokenizers perform suboptimally in conversational contexts, one of the most prevalent use cases of LLMs. By retraining tokenizers on a representative dataset of chatbot dialogues, we observed consistent reductions in the number of tokens required, with savings ranging from 5\% to over 10\% depending on the model. These reductions can translate into meaningful computational and energy savings, particularly in large-scale deployments where inference dominates the energy footprint of AI systems.

Nonetheless, the tokenizer affects how information is represented and learned during pretraining, and conversation-optimized tokenizers may result in representations that are less effective.  Therefore, future work should carefully evaluate the impact on downstream performance to ensure that gains in inference efficiency do not come at the expense of model quality, generalization, or versatility.

Further research should also extend the analysis to a broader set of conversational corpora, as well as to other types of texts, to evaluate in more detail the performance of the tokenizers. Finally, incorporating tokenization considerations into the training process itself, rather than treating them as fixed preprocessing steps, could open new avenues for co-designing tokenizers and model architectures tailored to specific use cases.

%\clearpage

%\subsection{Appendices}

%Use \verb|\appendix| before any appendix section to switch the section numbering over to %letters. See Appendix~\ref{sec:appendix} for an example.

\section*{Limitations}

The work presented in this paper has the following limitations:

\begin{enumerate}
    \item Only one dataset is used for conversations, and only one training corpus. Additional and larger datasets should be evaluated to ensure that the results do not depend on the dataset. 
    \item The study only considers the inference phase when the LLM is used to generate text. To have a comprehensive analysis of potential energy savings, the LLM training phase should also be considered.
    \item The potential impact of tokenization on LLM performance is not considered in the paper as it would require training the LLMs which is computationally unfeasible for a research group. However, previous work has shown that the impact is limited in many cases \cite{ali2023tokenizer}.
    \item The language distributions of the original tokenizers training corpora are unknown. A mismatch between these and the language distribution of the corpus used in this work to train the tokenizers may lead to some languages being favored while others are penalized.
    \item As with natural languages, programming code tokenization might be favored or penalized depending on each tokenizer’s original vocabulary and the amount of code in each programming language present in the conversations used to re-train them.

\end{enumerate}

%\section*{Acknowledgments}

\bibliography{custom}

%\appendix

%\section{Example Appendix}
%\label{sec:appendix}

%This is an appendix.

\end{document}